\documentclass{article}

\usepackage{PRIMEarxiv}

\usepackage[utf8]{inputenc}
\usepackage[T1]{fontenc}
\usepackage{hyperref}
\usepackage{url}
\usepackage{booktabs}
\usepackage{amsfonts}
\usepackage{amssymb}
\usepackage{amsmath}
\usepackage{mathtools}
\usepackage{nicefrac}
\usepackage{microtype}
\usepackage{graphicx}
\usepackage{multirow}
\usepackage{array}
\usepackage{longtable}
\usepackage{float}
\usepackage{subcaption}
\usepackage{natbib}
\usepackage[capitalize,noabbrev]{cleveref}

\DeclareMathOperator*{\argmax}{arg\,max}

\title{
    \includegraphics[width=\textwidth]{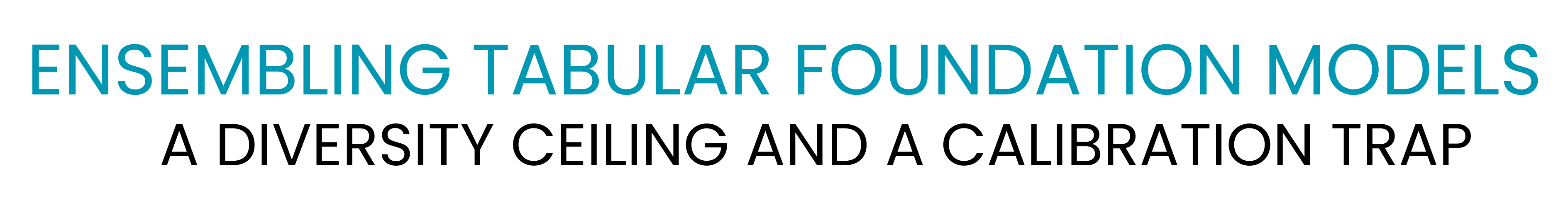}
}

\author{
  Aditya Tanna, Yash Desai, Pratinav Seth, \\
  Mohamed Bouadi, Nassim Bouarour, Vinay Kumar Sankarapu \\
  \affiliation{Lexsi Labs}
}

\setabstract{%
Tabular foundation models (TFMs) now match or beat tuned gradient-boosted trees on a growing fraction of tabular tasks, but no single TFM wins on every dataset. Ensembling is the go to fix here, and it works less well than expected. Six modern TFMs form a near-redundant pool: their mean pairwise Q-statistic is $0.961$, close enough to $1$ that any convex combination is bounded above. We benchmark six ensemble strategies over six TFMs on 153 OpenML classification tasks. The best ensemble, two-level cascade stacking, buys $+0.18\%$ accuracy over the strongest single TFM at $253\times$ the compute. A Friedman and Nemenyi analysis places three ensembles and the best base TFM in a single equivalence group; three other ensembles are significantly \emph{worse} than the best base. Stacking with a logistic-regression meta-learner is the most striking case: competitive accuracy and ROC-AUC, the worst log-loss rank among the ensembles. The meta-learner improves accuracy by sharpening class boundaries, which destroys calibration. We recommend greedy selection as the practical default.
}

\setkeywords{Tabular foundation models, ensembling, benchmarking, calibration, diversity}

\runningtitle{Ensembling Tabular Foundation Models}

\begin{document}
\maketitle

\section{Introduction}
\label{sec:intro}

Tabular foundation models (TFMs) have advanced rapidly, with growing architectural diversity. We show that this diversity is mostly nominal: a pool of six modern TFMs produces near-redundant predictions, and ensembling cannot exploit diversity that is not there. TabPFN \citep{hollmann2023tabpfn,hollmann2025tabpfnv2}, TabICL \citep{qu2025tabicl}, and variants like Mitra \citep{mitra2025}, Orion-Bix \citep{bouadi2025orionbix}, TabDPT \citep{tabdpt2024}, and CARTE \citep{kim2024carte} all perform in-context learning (ICL) over synthetic or curated priors. Continued pre-training on real-world tables \citep{garg2025realtabpfn} now rivals heavy AutoML in a single forward pass: TabPFNv2.5 matches AutoGluon~1.4 on TabArena \citep{erickson2026tabarena}.

What these models do not do, however, is win uniformly. Across our 153-dataset benchmark, the per-dataset accuracy wins among six TFMs split as follows: TabPFNv2.5-52, TabICLv2-46, TabPFNv2.6-25, TabICL-13, LimiX-12, OrionMSPv1.5-5 (all approximate numbers). The median accuracy spread between best and worst TFM on a single dataset is $1.95\%$; on $24\%$ of datasets it exceeds $5\%$, so committing to one TFM in advance loses meaningful accuracy on roughly a quarter of the benchmark. Ensembling is the textbook response \citep{breiman1996bagging,dietterich2000ensemble,caruana2004ensemble}.

The cost-benefit story for TFMs differs from the gradient-boosted decision tree (GBDT) setting where ensemble methods originated. Per-task TFM inference is cheap once the model is pretrained (under one second on most benchmark datasets), though pretraining itself is a substantial fixed cost. Cascade-style stacking layers $K$-fold OOF inference across all bases, dwarfing any single forward pass. Beyond compute, there is a structural concern. TFMs trained with ICL on synthetic priors approximate Bayesian model averaging at inference \citep{muller2022priorfit,zhang2023icl_bma}; broadly similar priors yield broadly similar posteriors, leaving little for a convex combiner to recover.

\paragraph{Position relative to existing work.} Recent work ensembles TFMs differently. The post-hoc ensemble protocol bundled with TabPFN \citep{hollmann2025tabpfnv2} ensembles hyperparameter configurations of \emph{one} TFM; TabICLv2 \citep{qu2025tabicl} does the same internally with column and class shuffles. TabArena \citep{erickson2026tabarena} measures post-hoc ensembling across \emph{heterogeneous} model classes (TFMs, GBDTs, neural baselines) and reports that some classes are over-represented in cross-class ensembles due to validation overfitting. HAPEns \citep{maier2026hapens} adds hardware-aware multi-objective selection on a general model pool. TabM \citep{gorishniy2025tabm} pursues parameter-efficient ensembling at the architecture level. Recent architectural work in the same TFM family includes Orion-BiX \citep{bouadi2025orionbix}, while \cite{tanna2026finetuning} study fine-tuning protocols (zero-shot, meta-learning, SFT, PEFT) across the same CC18/TALENT/TabZilla benchmarks we use here. We isolate a different question: holding the model \emph{class} fixed, what does ensembling six different TFMs buy on its own? Section~\ref{sec:related} situates the answer against the broader ensemble and tabular-deep-learning literature.

\paragraph{Contributions.}
\begin{enumerate}
    \item A diversity-ceiling diagnostic. Six modern TFMs share an ICL-on-synthetic-priors recipe and produce near-redundant predictions ($Q=0.961$), bounding the gain available to any convex combiner.
    \item An empirical accuracy/compute frontier with a calibration overlay across six ensemble strategies on 153 OpenML tasks. The best ensemble buys a sub-percent accuracy gain at $253\times$ the compute of the strongest base; three of six ensembles are significantly worse than the best base.
    \item Calibration findings. Stacking with a logistic-regression meta-learner has competitive accuracy and ROC-AUC ranks but the worst log-loss rank among the ensembles; the meta-learner sharpens class boundaries, improving accuracy at the cost of probability quality.
\end{enumerate}

\section{Related Work}
\label{sec:related}

The ensemble literature provides a useful baseline for what to expect from this study. Classical results show that gains from averaging classifiers scale with two things: how accurate the base classifiers are, and how independent their errors are. Bagging \citep{breiman1996bagging} reduces variance when bases are decorrelated; stacking \citep{wolpert1992stacked,ting1999issues} learns a meta-mapping over base outputs; ensemble selection from a large library of fits \citep{caruana2004ensemble} chooses a small weighted subset by greedy validation search. Across all three, the standard assumption is that the pool was constructed to be diverse, often by training the same model class on resampled or perturbed data. Diversity measures such as the Q-statistic, Cohen's $\kappa$, and disagreement \citep{kuncheva2003diversity} were introduced to quantify exactly that.

Tabular foundation models break the assumption. The bases are pretrained, not refit per dataset, and the per-task perturbation budget is small: random seeds change the in-context order but not the underlying prior. Recent work has explored ensembling in this setting at three different levels. At the architecture level, TabM \citep{gorishniy2025tabm} shares parameters across an internal set of branches and trains them end-to-end. At the configuration level, the post-hoc protocol in TabPFN \citep{hollmann2025tabpfnv2} averages many hyperparameter configurations of a single model. At the cross-class level, TabArena \citep{erickson2026tabarena} reports post-hoc ensembles over TFMs, GBDTs, and neural baselines, and notes that validation-based weight selection can over-represent some model classes. HAPEns \citep{maier2026hapens} extends that idea with a hardware-aware multi-objective selector. We sit at a fourth level: holding the model class fixed (all TFMs), and asking what convex or stacked combinations of six different pretrained TFMs can recover on their own.

The notion that ensemble gains have a ceiling is not new \citep{kuncheva2003diversity,dietterich2000ensemble}, but it usually appears as a property of small classifier pools on small datasets, not as a structural feature of a pretraining family. For TFMs, the question becomes whether the inductive bias of ICL over synthetic priors leaves enough room for diversity to matter. There is theoretical reason to expect that it does not: \cite{muller2022priorfit} and \cite{zhang2023icl_bma} formalise the sense in which ICL approximates Bayesian model averaging at inference time, so two TFMs trained on broadly similar priors are already implicitly averaging over broadly similar posterior families. The Q-statistic we report ($\bar{Q}=0.961$) is the empirical counterpart of that argument: six models that all consult the same kind of prior fail on the same instances. Real-TabPFN \citep{garg2025realtabpfn} suggests one possible escape route, namely continued pre-training on real-world tables to shift the prior; whether that produces enough error decorrelation to lift the ceiling is open.

Calibration is the second thread. Modern deep classifiers tend to be overconfident, and post-hoc fixes such as temperature scaling \citep{guo2017calibration} were developed specifically for that regime. \cite{niculescuMizil2005predicting} showed that some ensemble methods (notably bagged trees and random forests) produce well-calibrated probabilities almost as a byproduct, while others (boosting) do not. The pattern we find in \S\ref{sec:calibration} sits in the same family of results: convex combiners of TFM probabilities preserve the calibration of their inputs, but a discriminative meta-learner trained on out-of-fold predictions does not. Selective-prediction and worst-group metrics \citep{geifman2017selective,sagawa2019distributionally} let us tell the two failure modes apart, which is why we report them alongside accuracy.

\section{Ensemble Strategies}
\label{sec:strategies}

Six strategies share a common \texttt{fit}/\allowbreak\texttt{predict}/\allowbreak\texttt{predict\_proba} interface over a fixed pool of $K$ base TFMs producing class-probability vectors $p_k(x)$.

\textbf{Weighted Averaging (WA)} $\hat p = \sum_k w_k p_k$ with $w_k \propto \mathrm{score}_k$ on validation. No second stage. Cheapest combiner.

\textbf{Greedy Selection} \citep{caruana2004ensemble}. Forward selection with replacement: at each of $S$ iterations, the base whose addition maximises validation accuracy is added. Final weight equals selection count over $S$. We use $S=50$, matching the AutoGluon \citep{erickson2020autogluon} \texttt{WeightedEnsembleModel} default.

\textbf{Stacking} \citep{wolpert1992stacked,ting1999issues}. Bases produce $5$-fold out-of-fold (OOF) predictions. A logistic-regression meta-learner is trained on the OOF features.

\textbf{Temperature-Scaled Blending} \citep{guo2017calibration}. Per-base temperature $T_k$ is fit on the validation set by minimising negative log-likelihood (NLL) of $\mathrm{softmax}(\log p_k / T_k)$; calibrated probabilities are then averaged uniformly.

\textbf{Cascade Stacking.} Two-level stacking with skip connections, modifying AutoGluon's high-quality preset \citep{erickson2020autogluon}. Level-1 OOF predictions concatenate with raw features and feed level-2 base models, also with $K$-fold OOF. A final greedy-selection layer combines all level outputs. We use $2$ levels, $3$-fold OOF, $S=50$.

\textbf{Random-Init (Deep) Ensemble} \citep{lakshminarayanan2017simple}. Each TFM is run with $M=3$ different seeds. Per-base predictions are averaged across seeds, then cross-base averaging uses performance weights.

Strategies are implemented in Python on top of the public TabTune library \citep{tanna2025tabtune}.

\section{Experimental Setup}
\label{sec:setup}

\textbf{Datasets.} 153 OpenML classification tasks drawn from the CC18 \citep{bischl2017openml}, TALENT \citep{JMLR:v26:25-0512}, and TabZilla \citep{mcelfresh2023tabzilla} pools. Selection criteria and the full dataset inventory are in Appendix~\ref{app:dataset-details}.

\textbf{Base TFMs.} Six models in inference mode: TabPFNv2.5 \citep{Grinsztajn2025TabPFN25AT}, TabPFNv2.6 \citep{hollmann2025tabpfnv2}, TabICL \citep{qu2025tabicl}, TabICLv2 \citep{qu2025tabicl}, LimiX \citep{zhang2025limix}, OrionMSPv1.5 \citep{bouadi2025orionmsp}.

\textbf{Protocol.} Per dataset: an $80/20$ stratified train/test split; within train, a $75/25$ train/validation split for ensemble weight learning. Stacking and cascade levels use $5$-fold and $3$-fold internal CV respectively. A fixed seed controls splits and base-model initialisation.

\textbf{Metrics.} Accuracy, weighted F1, one-vs-rest ROC-AUC, multi-class log-loss, and total fit-time per dataset (seconds). For deeper analysis on the TabArena classification suite we additionally report expected calibration error (ECE) \citep{guo2017calibration}, the reliability component of the Brier decomposition \citep{brier1950verification}, area under the risk-coverage curve (AURC) \citep{geifman2017selective}, coverage at $95\%$ accuracy, and worst-group accuracy (WGA) \citep{sagawa2019distributionally}. Statistical significance is reported via Friedman \citep{friedman1940comparison}, Nemenyi \citep{nemenyi1963distribution}, and pairwise Wilcoxon signed-rank \citep{wilcoxon1945signed}.

\textbf{Hardware.} A single H100 (80\,GB) GPU per run.

\section{Results}
\label{sec:results}

\subsection{Aggregate performance}

Table~\ref{tab:main} reports per-method statistics across the 153 datasets in our benchmark. The accuracy spread among the top eight methods is $0.45$ percentage points (TabICL at $0.872$ to Cascade at $0.882$). The Friedman test rejects equality of mean ranks across the 12 methods ($\chi^2 = 389.95$, $p < 10^{-30}$); methods are not exchangeable, but the question is which differences survive a per-pair correction. Calibration, selective-prediction, and group-robustness metrics on the TabArena suite are reported in Table~\ref{tab:ablation} (Appendix~\ref{app:calibration}) and analysed in \S\ref{sec:calibration}.

\begin{table}[pt]
\centering
\small
\caption{Mean performance over 153 OpenML classification tasks. Lower rank is better. Methods are sorted by accuracy rank; ensemble methods are italicized. Best in each rank column is bold.}
\label{tab:main}
\begin{tabular}{lrrrrrrr}
\toprule
\textbf{Method} & \textbf{Acc.} & \textbf{Acc.\ rank} & \textbf{Log-loss} & \textbf{Log-loss rank} & \textbf{ROC-AUC} & \textbf{ROC-AUC rank} & \textbf{Fit (s)} \\
\midrule
\textit{Cascade\_2level}     & 0.882 & \textbf{4.48} & 0.289 & 4.55          & 0.907 & 5.01          & 178.5 \\
\textit{Stacking\_LR}        & 0.880 & 4.96          & 0.288 & 8.13          & 0.907 & \textbf{4.59} & 6.6   \\
\textit{Greedy\_Selection}   & 0.881 & 5.28          & 0.278 & 4.54          & 0.906 & 5.51          & 6.7   \\
TabICLv2                     & 0.881 & 5.39          & 0.274 & \textbf{4.11} & 0.907 & 4.73          & \textbf{0.7} \\
\textit{DeepEnsemble\_3seed} & 0.879 & 6.10          & 0.281 & 5.62          & 0.907 & 5.58          & 75.7  \\
\textit{WA\_performance}     & 0.879 & 6.18          & 0.282 & 5.50          & 0.907 & 5.49          & 6.6   \\
\textit{Temp\_Scaled}        & 0.879 & 6.41          & 0.283 & 6.47          & 0.907 & 6.14          & 6.6   \\
TabPFNv2.5                   & 0.878 & 6.44          & 0.286 & 6.64          & 0.904 & 7.69          & 1.2   \\
TabPFNv2.6                   & 0.876 & 6.78          & 0.283 & 6.26          & 0.903 & 6.89          & 1.6   \\
LimiX                        & 0.875 & 7.67          & 0.288 & 7.17          & 0.902 & 7.64          & 0.5   \\
TabICL                       & 0.872 & 7.96          & 0.289 & 7.61          & 0.900 & 8.02          & 1.0   \\
OrionMSPv1.5                 & 0.848 & 10.34         & 0.356 & 11.41         & 0.880 & 10.72         & 1.8   \\
\bottomrule
\end{tabular}
\end{table}

The Nemenyi critical difference at $\alpha=0.05$ for $K=12, N=153$ is $\mathrm{CD}=1.347$. Three methods sit within CD of the top-ranked Cascade\_2level: Stacking\_LR ($\Delta=0.48$), Greedy\_Selection ($\Delta=0.80$), and TabICLv2 ($\Delta=0.90$). Three ensembles and one base TFM are statistically indistinguishable on accuracy across 153 tasks; the remaining three ensembles cannot beat the best base. Pairwise Wilcoxon tests sharpen the picture: against TabICLv2, only Cascade\_2level wins ($+0.18\%$, $p=0.008$); Greedy\_Selection ($+0.01\%$) and Stacking\_LR ($-0.03\%$) tie; WA, Temp\_Scaled, and DeepEnsemble\_3seed are all significantly worse ($p<0.05$). One ensemble of six clears the bar of beating the strongest base.

\subsection{Accuracy/compute frontier}

\begin{figure}[pt]
\centering
\includegraphics[width=0.63\textwidth]{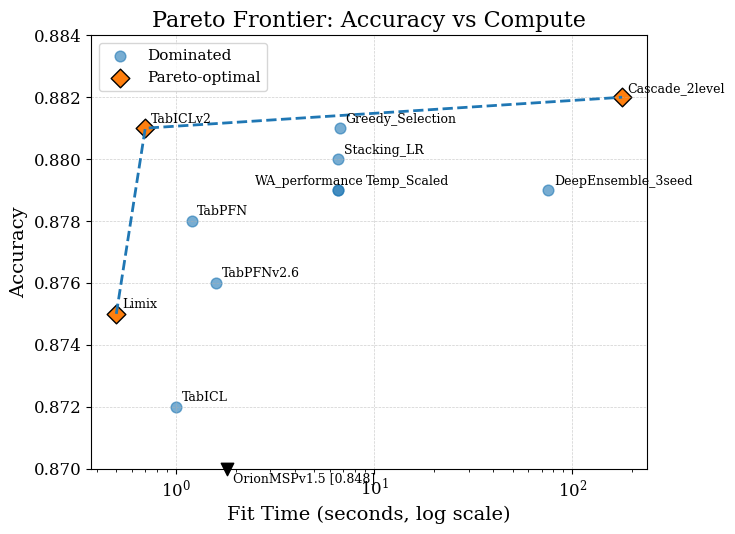}
\caption{Mean accuracy vs.\ mean fit time (log scale, seconds per dataset) over 153 OpenML tasks. Pareto-optimal methods are marked. Cascade buys an accuracy advantage over TabICLv2 for $253\times$ the compute.}
\label{fig:pareto}
\end{figure}

Fit times span four orders of magnitude. TabICLv2 averages $0.71$\,s per dataset; Greedy, Stacking\_LR, WA, and Temp\_Scaled all sit near $6.6$\,s, which is roughly the cost of one forward pass through the six bases plus a thin combination layer. DeepEnsemble\_3seed costs $75.7$\,s, and Cascade\_2level costs $178.5$\,s. Figure~\ref{fig:pareto} plots the trade-off.

The Pareto frontier is dominated by TabICLv2 (cheapest competitive option) and Greedy\_Selection (best accuracy at moderate cost). Cascade\_2level sits on the frontier, but its marginal accuracy advantage over TabICLv2 corresponds to a $253\times$ compute multiplier. DeepEnsemble\_3seed is dominated outright: WA\_performance and Stacking\_LR achieve similar or better accuracy at one tenth its cost. Figure~\ref{fig:cd} (Appendix~\ref{app:cd}) shows the corresponding critical-difference diagram.

\subsection{Calibration and the diversity ceiling}
\label{sec:calibration}

\textbf{Log-loss diverges from accuracy.} The log-loss column of Table~\ref{tab:main} tells a different story than the accuracy column. TabICLv2 has the lowest log-loss rank ($4.11$); Greedy\_Selection ($4.54$) and Cascade\_2level ($4.55$) sit close behind, both producing convex combinations of probability vectors. Stacking\_LR ranks $8.13$, the worst of any method tested except OrionMSPv1.5. Linear stacking still places the right class label, which is why its accuracy and ROC-AUC ranks stay competitive, but the cross-entropy objective on OOF predictions pushes the meta-learner toward sharper probability outputs than the bases produce, which degrades calibration.

\textbf{Calibration tracks combination strategy, not compute.} Table~\ref{tab:ablation} (Appendix~\ref{app:calibration}) reports five complementary metrics on the TabArena classification suite: ECE, Brier reliability, AURC, coverage at $95\%$ accuracy, and worst-group accuracy. TabICLv2 sets the calibration ceiling (ECE $=0.0236$, Brier-REL $=0.0024$). Greedy\_Selection is the only ensemble that approaches it (ECE $=0.0253$), and it never optimises for calibration directly. Stacking\_LR records the worst calibration of any ensemble (ECE $=0.0272$, Brier-REL $=0.0031$), consistent with its log-loss rank. Temperature-Scaled Blending is equally instructive: per-base NLL minimisation gives an ECE of $0.0273$, no better than Stacking\_LR's.

\textbf{Base-model diversity caps uncertainty quality.} The mean pairwise Q-statistic \citep{kuncheva2003diversity} across the six TFMs is $0.961$ ($\sigma=0.183$, Cohen's $\kappa=0.856$, in the ``almost perfect agreement'' band of the conventional Landis-Koch scale). Q values close to $1$ signal near-redundancy: the six bases share the ICL-on-synthetic-priors recipe and tend to fail on the same instances, so any convex combiner has little variance to absorb; Appendix~\ref{app:proposition} states this ceiling formally as a consensus-set bound on the ensemble-vs-base accuracy gap. The ceiling is most consequential for DeepEnsemble\_3seed: three random seeds perturb context order but share the synthetic prior, producing an AURC of $0.0617$ ($28\%$ above TabICLv2's $0.0483$) and coverage at $95\%$ accuracy of $62.1\%$ versus $68.4\%$. The $75.7$\,s cost buys neither accuracy nor uncertainty improvement; for selective prediction, Greedy\_Selection (AURC $=0.0484$) is the appropriate choice.

\textbf{Cascade earns its cost on group robustness.} Worst-group accuracy is the one axis where heavy stacking earns its overhead. Cascade\_2level reaches $0.803$, on par with TabICLv2 ($0.802$) and outperforming all other ensembles by roughly three points (Greedy and Stacking\_LR both at $0.776$). The skip-connection architecture appears to implicitly down-weight base models that are systematically biased on minority subgroups; simpler convex combiners do not replicate this. The fairness margin is narrow and confined to sensitive-attribute datasets, but it is the one regime in which cascade's $253\times$ compute overhead translates into a qualitative advantage rather than a marginal one.

\textbf{Per-dataset patterns.} Aggregate means hide per-dataset behaviour. Cascade\_2level beats the per-task oracle base on $42$ of $153$ tasks (ties on $36$, loses on $75$, mean $\Delta=-0.0025$); against TabICLv2 specifically, $70$/$41$/$42$ (mean $\Delta=+0.0018$). Ensembles win on average not because they are uniformly better, but because no fixed base is best on every dataset. The Pearson correlation between ensemble gain and inter-base accuracy spread is $-0.03$, so the ceiling is roughly uniform across dataset shapes; ensembling helps no more on high-spread datasets than on low-spread ones.

\section{Discussion}
\label{sec:discussion}

The shape of the result holds across every metric we report. TFMs trained with ICL on synthetic priors already approximate Bayesian model averaging at inference, so explicit downstream ensembling lands inside the noise floor of a strong single base. Cascade stacking buys the last $0.2\%$ of accuracy by letting the meta-learner combine raw features alongside OOF predictions, but the cost-benefit ratio is poor outside competition settings. Greedy selection is the more honest default: roughly $10\times$ the cost of the strongest single model, the same mean accuracy as the heaviest stack, and no calibration regression.

Two patterns matter for downstream work. First, calibration is not a free byproduct of accuracy ensembling. Stacking with logistic regression damages probability quality despite improving accuracy rank, and uniform averaging after per-base temperature scaling is no better than the best base. Calibration-aware meta-learners optimised for log-loss directly, or post-hoc recalibration applied to the ensemble output rather than to its members, both remain open. Second, the per-dataset variance in best-base identity is what ensembling is really being asked to solve. A small gating learner trained on dataset metafeatures to pick the best TFM per dataset might match cascade at far lower compute, and is a better fit to the structure of the problem than a stack.

\subsection{Limitations.} 
We use a single seed per dataset; statistical power comes from across-dataset variation rather than within-dataset replicates, which is what paired Wilcoxon, Friedman, and Nemenyi tests assume on $N=153$ tasks. A small number of the largest tasks were dropped for individual base TFMs due to memory constraints, and Table~\ref{tab:main} reports the intersection of tasks that completed for every method. We do not include GBDT baselines; the comparison is between single TFMs and TFM-only ensembles, leaving open whether out-of-class diversity (TFM + GBDT) recovers gains the within-class pool cannot.

\subsection{Contamination.} 
Several base TFMs were pretrained on data overlapping with OpenML, a concern raised by the TabArena protocol \citep{erickson2026tabarena}. Reported deltas should be read as upper bounds on within-pool ensemble effects: under cleaner contamination protocols, the ensemble-vs-best-base gap is likely smaller, not larger. Contamination does not change the qualitative finding (a near-redundant pool produces a hard ceiling), but the precise accuracy delta is best treated as inflated.

\textbf{Future work.} Hybrid TFM+GBDT pools, per-dataset gating learners, time-series TFMs where base-model spread may be larger, and calibration-aware meta-learners.

\section{Conclusion}
\label{sec:conclusion}

Six tabular foundation models trained with ICL on synthetic priors form a near-redundant pool. A Q-statistic of $0.961$ caps what any convex combiner can recover, and the empirical results follow: a top equivalence group of four methods (three ensembles plus the best base) statistically indistinguishable on accuracy; a best ensemble that recovers a sub-percent accuracy gain at $253\times$ compute; and a calibration trap when a meta-learner is asked to manufacture extra accuracy by sharpening probabilities. Greedy selection is the practical default; cascade stacking is justifiable only when worst-group accuracy is a primary target. \textit{The open question is whether out-of-class diversity (TFM + GBDT) breaks through the ceiling that within-class pools cannot.}

\bibliographystyle{unsrt}
\bibliography{references}

\appendix
\clearpage
\section{Method-name glossary}
\label{app:glossary}

The body and Tables~\ref{tab:main} and~\ref{tab:ablation} use compact short-form labels; Figures~\ref{fig:h2h_win_matrix} and~\ref{fig:rank} render the same methods in long form. Table~\ref{tab:glossary} reconciles the two.

\begin{table}[H]
\centering
\small
\caption{Short-form labels used in prose and Tables~\ref{tab:main} and~\ref{tab:ablation} mapped to the long-form labels rendered in Figures~\ref{fig:h2h_win_matrix} and~\ref{fig:rank}.}
\label{tab:glossary}
\begin{tabular}{ll}
\toprule
\textbf{Short form (prose, tables)} & \textbf{Long form (figures)} \\
\midrule
\texttt{Cascade\_2level}      & Cascade 2-Level Stacking \\
\texttt{Stacking\_LR}         & Logistic Regression Stacking \\
\texttt{Greedy\_Selection}    & Greedy Selection \\
\texttt{WA\_performance}      & Weighted Averaging \\
\texttt{Temp\_Scaled}         & Temperature Scaling \\
\texttt{DeepEnsemble\_3seed}  & Deep Ensemble \\
\bottomrule
\end{tabular}
\end{table}
\clearpage
\section{Critical-difference diagram}
\label{app:cd}

\begin{figure}[H]
\centering
\includegraphics[width=0.7\textwidth]{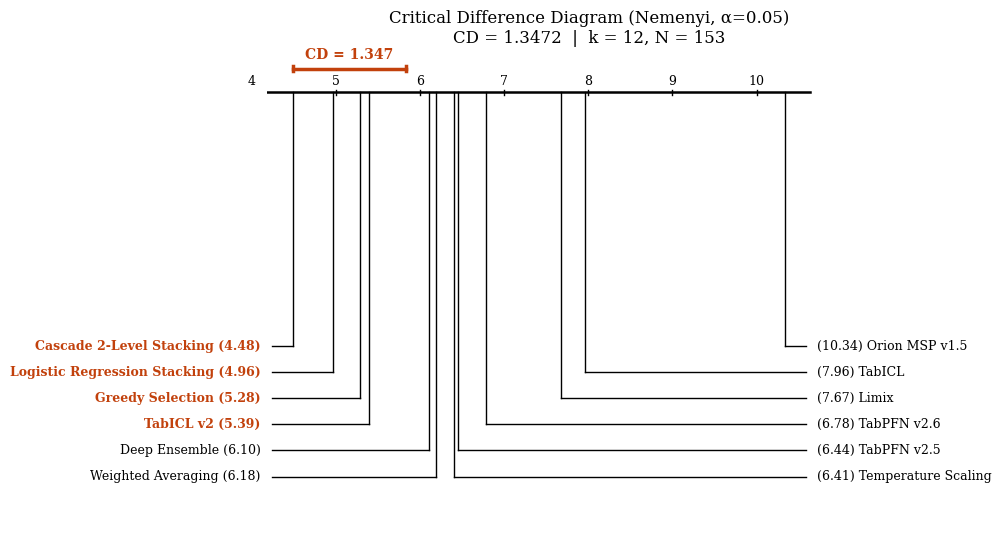}
\caption{Critical-difference diagram for accuracy ranks (Friedman + Nemenyi, $\alpha=0.05$, $\mathrm{CD}=1.347$). Methods connected by horizontal bars are statistically indistinguishable. The top equivalence group is Cascade\_2level, Stacking\_LR, Greedy\_Selection, and TabICLv2.}
\label{fig:cd}
\end{figure}

Figure~\ref{fig:cd} visualises the rank-difference structure underlying the Friedman and Nemenyi analysis in \S\ref{sec:results}. Methods are positioned along the mean-rank axis; horizontal bars connect groups whose pairwise differences fall within the critical difference $\mathrm{CD}=1.347$ (so members of a group are statistically indistinguishable on accuracy at $\alpha=0.05$).

\clearpage
\section{Calibration, uncertainty, and group-robustness metrics}
\label{app:calibration}

\begin{table}[H]
\centering
\small
\caption{Calibration, uncertainty, and group-robustness metrics on the TabArena classification suite. ECE: expected calibration error (15-bin). Brier-REL: reliability component of the Brier decomposition. AURC: area under the risk-coverage curve (max-probability scorer). Cov@95: fraction of samples predictable at $\geq 95\%$ accuracy. WGA: worst-group accuracy on sensitive-attribute datasets. Best ensemble and best base are bolded separately.}
\label{tab:ablation}
\begin{tabular}{lrrrrr}
\toprule
\textbf{Method} & \textbf{ECE$\downarrow$} & \textbf{Brier-REL$\downarrow$} & \textbf{AURC$\downarrow$} & \textbf{Cov@95$\uparrow$} & \textbf{WGA$\uparrow$} \\
\midrule
\textit{Cascade\_2level}     & 0.0261          & \textbf{0.0025} & 0.0485          & 0.677          & \textbf{0.803} \\
\textit{Greedy\_Selection}   & \textbf{0.0253} & 0.0027          & \textbf{0.0484} & 0.679          & 0.776 \\
\textit{Stacking\_LR}        & 0.0272          & 0.0031          & 0.0491          & \textbf{0.684} & 0.776 \\
\textit{Temp\_Scaled}        & 0.0273          & 0.0030          & 0.0485          & 0.679          & 0.790 \\
\textit{WA\_performance}     & 0.0277          & 0.0030          & 0.0485          & 0.679          & 0.790 \\
\textit{DeepEnsemble\_3seed} & 0.0275          & 0.0030          & 0.0617          & 0.621          & 0.790 \\
\midrule
TabICLv2                     & \textbf{0.0236} & \textbf{0.0024} & \textbf{0.0483} & \textbf{0.684} & \textbf{0.802} \\
TabICL                       & 0.0261          & 0.0028          & 0.0498          & 0.673          & 0.775 \\
TabPFNv2.6                   & 0.0255          & 0.0028          & 0.0485          & 0.680          & 0.773 \\
TabPFNv2.5                   & 0.0258          & 0.0026          & 0.0490          & 0.670          & 0.779 \\
LimiX                        & 0.0252          & 0.0027          & 0.0496          & 0.673          & 0.789 \\
OrionMSPv1.5                 & 0.0419          & 0.0044          & 0.0603          & 0.629          & 0.758 \\
\bottomrule
\end{tabular}
\end{table}
\clearpage
\section{Convex-combination ceiling: a formal statement}
\label{app:proposition}

\textbf{Proposition (Convex-combination ceiling).} 

Let $f_1, \ldots, f_K$ be classifiers producing class-probability vectors $p_k(x)$ with hard predictions $\hat{y}_k(x) = \argmax_c p_k(x)_c$, and let $\mathcal{C} = \{x : \hat{y}_1(x) = \cdots = \hat{y}_K(x)\}$ denote the consensus set. For any convex combiner $\hat{p}(x) = \sum_k w_k p_k(x)$ with $w_k \geq 0$ and $\sum_k w_k = 1$, the ensemble prediction $\argmax_c \hat{p}(x)_c$ equals the unanimous label on $\mathcal{C}$. Consequently the accuracy gap between the ensemble and any single base classifier is bounded above by the disagreement-set fraction $1 - |\mathcal{C}|/|\mathcal{X}|$.

\emph{Proof sketch.} On $x \in \mathcal{C}$ every $p_k(x)$ assigns its maximum to the same label $y^*$, so $\hat{p}(x)_{y^*} = \sum_k w_k p_k(x)_{y^*} \geq \sum_k w_k p_k(x)_c$ for every other class $c$, with equality only if every base ties at $x$. Hence $\hat{y}(x) = y^*$, matching every base on $\mathcal{C}$. Outside $\mathcal{C}$ the ensemble can differ from a base on at most $|\mathcal{X} \setminus \mathcal{C}|$ inputs, bounding the absolute accuracy gap.\hfill$\square$

A high mean pairwise Q-statistic is a sufficient indicator that, when individual error rates are similar across the pool, the disagreement set is small \citep{kuncheva2003diversity}: errors concentrate on the same instances, so the consensus set is large and the bound above tightens. For our six-TFM pool the mean pairwise Q-statistic is $\bar{Q} = 0.961$ (computation in Appendix~\ref{app:qstat}), close to $1$ and signalling a hard diversity ceiling any combiner must work against.

\section{Q-statistic computation}
\label{app:qstat}

The pairwise Q-statistic \citep{kuncheva2003diversity} is computed from each pair of base TFMs' test-set predictions on each of the 153 tasks. For a pair $(f_k, f_l)$ on a task, we form the $2 \times 2$ contingency table on test instances by (correct, wrong) under each classifier and define
\begin{equation*}
Q_{kl}^{\text{(task)}} = \frac{ad - bc}{ad + bc},
\end{equation*}
where $a$, $b$, $c$, $d$ are the counts of (both correct), ($k$ correct, $l$ wrong), ($k$ wrong, $l$ correct), and (both wrong) respectively. Per-pair $Q_{kl}$ is the unweighted mean of $Q_{kl}^{\text{(task)}}$ across the 153 tasks. The pool-level $\bar{Q} = 0.961$ reported in \S\ref{sec:calibration} is the unweighted mean across the $\binom{6}{2} = 15$ unordered base-pair indices; $\sigma = 0.183$ is the standard deviation across the same set. Cohen's $\kappa$ uses the same contingency tables and is reported using the standard chance-adjusted-agreement formula.

\clearpage
\section{Head-to-head ensemble comparison}
\label{app:h2h}

\begin{figure}[H]
\centering
\includegraphics[width=0.85\textwidth]{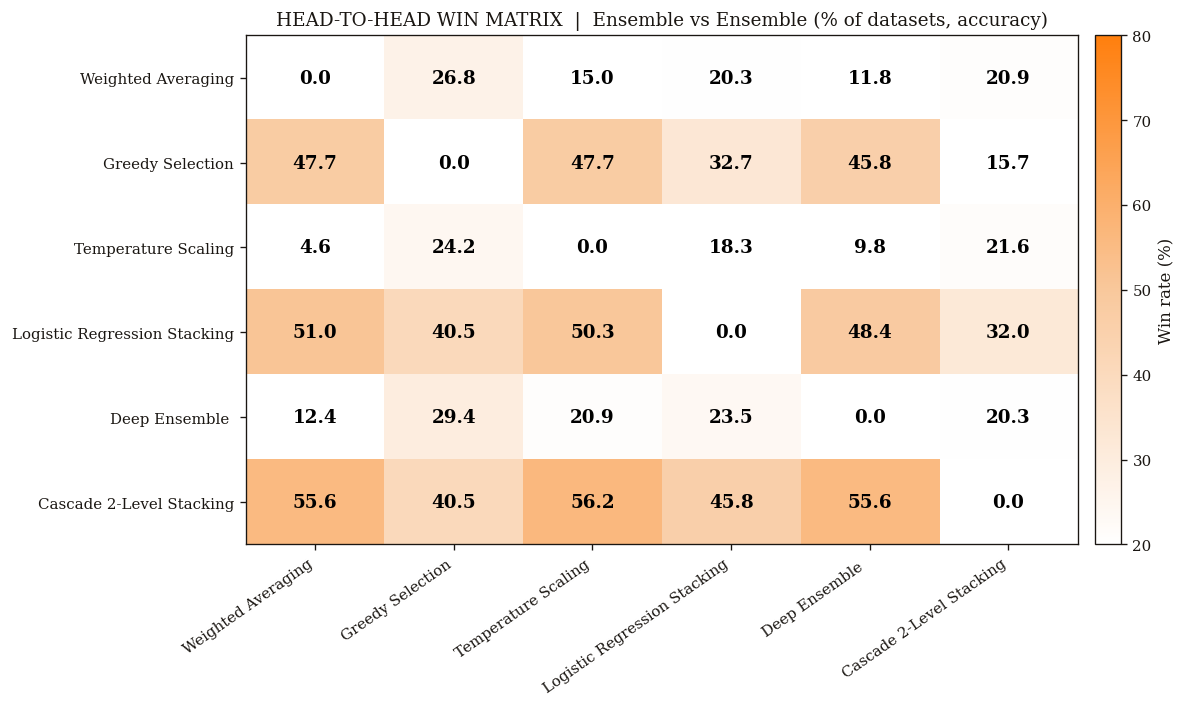}
\caption{Head-to-head win matrix across 153 OpenML classification tasks (accuracy). Each cell $(i,j)$ reports the percentage of datasets on which row method $i$ outperforms column method $j$; the diagonal is zero by construction. Cascade 2-Level Stacking is the dominant strategy ($55.6\%$, $56.2\%$, and $55.6\%$ of datasets against Weighted Averaging, Temperature Scaling, and Deep Ensemble respectively). Logistic Regression Stacking ranks second. Weighted Averaging wins on at most $26.8\%$ of datasets against any single opponent; Temperature Scaling never exceeds $24.2\%$.}
\label{fig:h2h_win_matrix}
\end{figure}

Figure~\ref{fig:h2h_win_matrix} reports per-pair win rates across the 153 tasks for the six ensemble methods. Cascade 2-Level Stacking is the dominant strategy: it wins on $55.6\%$, $56.2\%$, and $55.6\%$ of datasets against Weighted Averaging, Temperature Scaling, and Deep Ensemble respectively, and ties or loses only against Logistic Regression Stacking and Greedy Selection. Weighted Averaging never exceeds a $26.8\%$ win rate against any single opponent and Temperature Scaling never exceeds $24.2\%$, consistent with their ranking in Table~\ref{tab:main}.

\clearpage
\section{Mean rank leaderboard}
\label{app:rank}

\begin{figure}[H]
\centering
\includegraphics[width=0.85\textwidth]{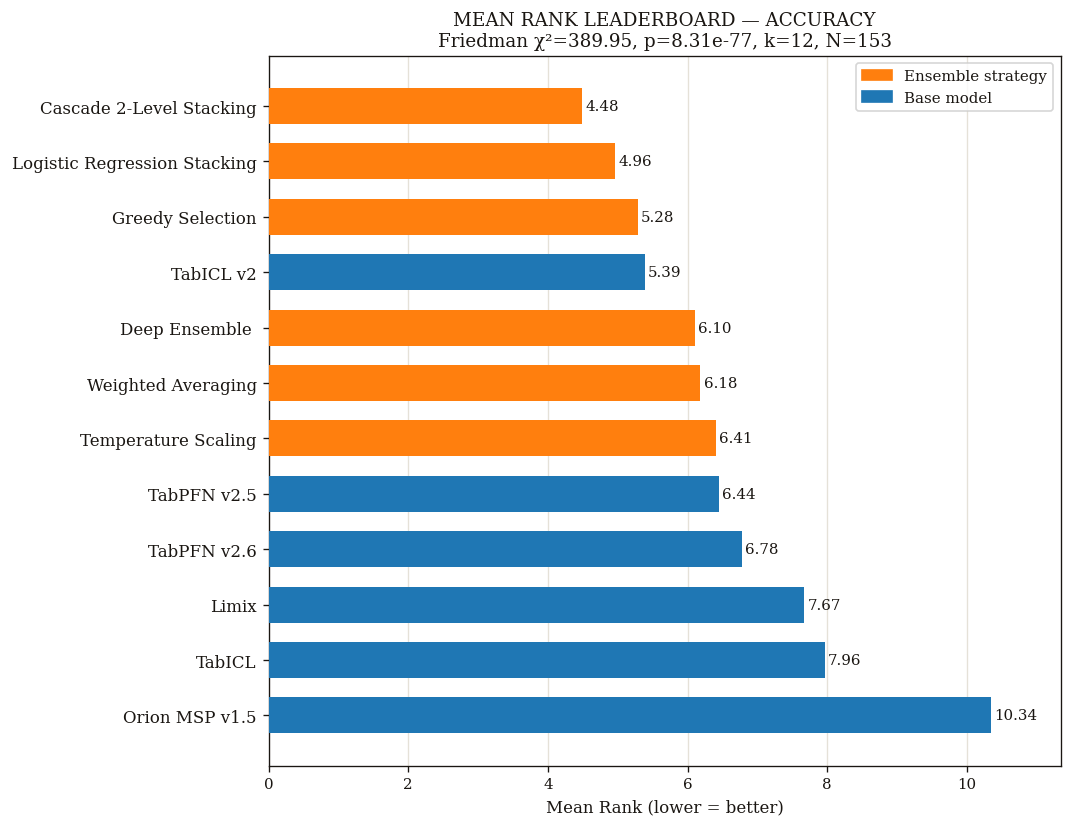}
\caption{Mean rank leaderboard for the 12 methods evaluated across 153 datasets. Ensemble strategies (orange) cluster with the strongest base models (blue) at the top of the ranking; Cascade 2-Level Stacking achieves the best rank ($4.48$), followed by Logistic Regression Stacking ($4.96$) and Greedy Selection ($5.28$). Friedman $\chi^2 = 389.95$, $p \approx 8.31\times10^{-77}$.}
\label{fig:rank}
\end{figure}

Figure~\ref{fig:rank} plots the per-method mean accuracy rank across the 153 tasks, colour-coded by method type (ensemble strategies in orange, base TFMs in blue). Cascade 2-Level Stacking holds the best mean rank ($4.48$), followed by Logistic Regression Stacking ($4.96$), Greedy Selection ($5.28$), and the strongest base TabICLv2 ($5.39$). The Friedman test rejects equality of mean ranks across the 12 methods at $\chi^2=389.95$, $p\approx 8.31\times 10^{-77}$; which pairwise differences survive the per-pair correction is shown in Figure~\ref{fig:cd}.

\clearpage

\section{Dataset details}
\label{app:dataset-details}

This section lists the full inventory of OpenML benchmark datasets used in our study. Table~\ref{tab:datasets} reports the OpenML identifier, dataset name, sample count, feature count, number of target classes, and source benchmark suite for each task.

\setlength{\LTcapwidth}{\textwidth}

\begingroup
\setlength{\tabcolsep}{5pt}
\renewcommand{\arraystretch}{1.0}

\begin{longtable}{@{\extracolsep{\fill}}r l r r r c@{}}
\caption{%
  Full inventory of the 153 OpenML benchmark datasets used in this study.
  \textbf{ID}: OpenML dataset identifier.
  \textbf{Samples}: number of instances (range: 128 to 581{,}012; median: 3{,}196).
  \textbf{Feat.}: number of input features (range: 5 to 1{,}777; median: 22).
  \textbf{Cls.}: number of target classes (range: 2 to 10; median: 2).
  \textbf{Source}: benchmark suite from which the task was drawn.
  \textbf{CC18}: OpenML-CC18; \textbf{TA}: TabArena; \textbf{TL}: Talent; \textbf{TZ}: TabZilla.
  Datasets shared across suites carry combined source tags (e.g., CC18,TZ).%
} \label{tab:datasets} \\
\toprule
\textbf{ID} & \textbf{Name} & \textbf{Samples} & \textbf{Feat.} & \textbf{Cls.} & \textbf{Source} \\
\midrule
\endfirsthead

\multicolumn{6}{c}{\tablename~\thetable\ \textit{(continued from previous page)}} \\
\toprule
\textbf{ID} & \textbf{Name} & \textbf{Samples} & \textbf{Feat.} & \textbf{Cls.} & \textbf{Source} \\
\midrule
\endhead

\midrule
\multicolumn{6}{r}{\textit{Continued on next page}} \\
\endfoot

\bottomrule
\endlastfoot

3     & \texttt{kr-vs-kp}                                                          & 3,196   & 37    & 2  & \texttt{CC18}     \\
11    & \texttt{balance-scale}                                                     & 625     & 5     & 3  & \texttt{CC18,TZ}  \\
12    & \texttt{mfeat-factors}                                                     & 2,000   & 217   & 10 & \texttt{CC18}     \\
14    & \texttt{mfeat-fourier}                                                     & 2,000   & 77    & 10 & \texttt{CC18,TZ}  \\
15    & \texttt{breast-w}                                                          & 699     & 10    & 2  & \texttt{CC18}     \\
16    & \texttt{mfeat-karhunen}                                                    & 2,000   & 65    & 10 & \texttt{CC18}     \\
18    & \texttt{mfeat-morphological}                                               & 2,000   & 7     & 10 & \texttt{CC18}     \\
21    & \texttt{car}                                                               & 1,728   & 7     & 4  & \texttt{TL}       \\
22    & \texttt{mfeat-zernike}                                                     & 2,000   & 48    & 10 & \texttt{CC18,TZ}  \\
23    & \texttt{cmc}                                                               & 1,473   & 10    & 3  & \texttt{CC18}     \\
27    & \texttt{colic}                                                             & 368     & 23    & 2  & \texttt{TZ}       \\
28    & \texttt{optdigits}                                                         & 5,620   & 65    & 10 & \texttt{CC18}     \\
29    & \texttt{credit-approval}                                                   & 690     & 16    & 2  & \texttt{CC18,TZ}  \\
30    & \texttt{page-blocks}                                                       & 5,473   & 11    & 5  & \texttt{TL}       \\
31    & \texttt{credit-g}                                                          & 1,000   & 21    & 2  & \texttt{CC18,TZ}  \\
32    & \texttt{pendigits}                                                         & 10,992  & 17    & 10 & \texttt{CC18}     \\
36    & \texttt{segment}                                                           & 2,310   & 20    & 7  & \texttt{TL}       \\
37    & \texttt{diabetes}                                                          & 768     & 9     & 2  & \texttt{CC18}     \\
38    & \texttt{sick}                                                              & 3,772   & 30    & 2  & \texttt{CC18}     \\
44    & \texttt{spambase}                                                          & 4,601   & 58    & 2  & \texttt{CC18}     \\
46    & \texttt{splice}                                                            & 3,190   & 61    & 3  & \texttt{CC18,TZ}  \\
50    & \texttt{tic-tac-toe}                                                       & 958     & 10    & 2  & \texttt{CC18}     \\
54    & \texttt{vehicle}                                                           & 846     & 19    & 4  & \texttt{CC18,TZ}  \\
60    & \texttt{waveform-5000}                                                     & 5,000   & 41    & 3  & \texttt{TL}       \\
151   & \texttt{electricity}                                                       & 45,312  & 9     & 2  & \texttt{CC18}     \\
179   & \texttt{adult}                                                             & 48,842  & 15    & 2  & \texttt{TL}       \\
180   & \texttt{covertype}                                                         & 110,393 & 55    & 7  & \texttt{TL}       \\
181   & \texttt{yeast}                                                             & 1,484   & 9     & 10 & \texttt{TL}       \\
182   & \texttt{satimage}                                                          & 6,430   & 37    & 6  & \texttt{CC18}     \\
188   & \texttt{eucalyptus}                                                        & 736     & 20    & 5  & \texttt{CC18}     \\
293   & \texttt{covertype}                                                         & 581,012 & 55    & 2  & \texttt{TL}       \\
333   & \texttt{monks-problems-1}                                                  & 556     & 7     & 2  & \texttt{TZ}       \\
458   & \texttt{analcatdata\_authorship}                                           & 841     & 71    & 4  & \texttt{CC18}     \\
469   & \texttt{analcatdata\_dmft}                                                 & 797     & 5     & 6  & \texttt{CC18}     \\
470   & \texttt{profb}                                                             & 672     & 10    & 2  & \texttt{TZ}       \\
554   & \texttt{mnist\_784}                                                        & 70,000  & 785   & 10 & \texttt{CC18}     \\
846   & \texttt{elevators}                                                         & 16,599  & 19    & 2  & \texttt{TZ}       \\
934   & \texttt{socmob}                                                            & 1,156   & 6     & 2  & \texttt{TZ}       \\
999   & \texttt{audiology}                                                         & 226     & 70    & 2  & \texttt{TZ}       \\
1038  & \texttt{gina\_agnostic}                                                    & 3,468   & 971   & 2  & \texttt{TL}       \\
1043  & \texttt{ada\_agnostic}                                                     & 4,562   & 49    & 2  & \texttt{TZ}       \\
1046  & \texttt{mozilla4}                                                          & 15,545  & 6     & 2  & \texttt{TL}       \\
1049  & \texttt{pc4}                                                               & 1,458   & 38    & 2  & \texttt{CC18}     \\
1050  & \texttt{pc3}                                                               & 1,563   & 38    & 2  & \texttt{CC18}     \\
1053  & \texttt{jm1}                                                               & 10,885  & 22    & 2  & \texttt{CC18}     \\
1063  & \texttt{kc2}                                                               & 522     & 22    & 2  & \texttt{CC18}     \\
1067  & \texttt{kc1}                                                               & 2,109   & 22    & 2  & \texttt{CC18,TZ}  \\
1068  & \texttt{pc1}                                                               & 1,109   & 22    & 2  & \texttt{CC18}     \\
1111  & \texttt{KDDCup09\_appetency}                                               & 50,000  & 231   & 2  & \texttt{TL}       \\
1112  & \texttt{KDDCup09\_churn}                                                   & 50,000  & 231   & 2  & \texttt{TL}       \\
1114  & \texttt{KDDCup09\_upselling}                                               & 50,000  & 231   & 2  & \texttt{TL}       \\
1116  & \texttt{musk}                                                              & 6,598   & 168   & 2  & \texttt{TL}       \\
1119  & \texttt{adult-census}                                                      & 32,561  & 16    & 2  & \texttt{TL}       \\
1120  & \texttt{MagicTelescope}                                                    & 19,020  & 12    & 2  & \texttt{TL}       \\
1169  & \texttt{airlines}                                                          & 539,383 & 8     & 2  & \texttt{TZ}       \\
1459  & \texttt{artificial-characters}                                             & 10,218  & 8     & 10 & \texttt{TZ}       \\
1461  & \texttt{bank-marketing}                                                    & 45,211  & 17    & 2  & \texttt{CC18}     \\
1462  & \texttt{banknote-authentication}                                           & 1,372   & 5     & 2  & \texttt{CC18}     \\
1464  & \texttt{blood-transfusion-service-center}                                  & 748     & 5     & 2  & \texttt{CC18}     \\
1467  & \texttt{climate-model-simulation-crashes}                                  & 540     & 21    & 2  & \texttt{TL}       \\
1468  & \texttt{cnae-9}                                                            & 1,080   & 857   & 9  & \texttt{CC18,TZ}  \\
1471  & \texttt{eeg-eye-state}                                                     & 14,980  & 15    & 2  & \texttt{TL}       \\
1475  & \texttt{first-order-theorem-proving}                                       & 6,118   & 52    & 6  & \texttt{CC18}     \\
1476  & \texttt{gas-drift}                                                         & 13,910  & 129   & 6  & \texttt{TL}       \\
1478  & \texttt{har}                                                               & 10,299  & 562   & 6  & \texttt{CC18}     \\
1480  & \texttt{ilpd}                                                              & 583     & 11    & 2  & \texttt{CC18}     \\
1485  & \texttt{madelon}                                                           & 2,600   & 501   & 2  & \texttt{CC18}     \\
1486  & \texttt{nomao}                                                             & 34,465  & 119   & 2  & \texttt{CC18,TZ}  \\
1487  & \texttt{ozone-level-8hr}                                                   & 2,534   & 73    & 2  & \texttt{CC18}     \\
1489  & \texttt{phoneme}                                                           & 5,404   & 6     & 2  & \texttt{CC18}     \\
1494  & \texttt{qsar-biodeg}                                                       & 1,055   & 42    & 2  & \texttt{CC18,TZ}  \\
1497  & \texttt{wall-robot-navigation}                                             & 5,456   & 25    & 4  & \texttt{CC18}     \\
1501  & \texttt{semeion}                                                           & 1,593   & 257   & 10 & \texttt{CC18}     \\
1510  & \texttt{wdbc}                                                              & 569     & 31    & 2  & \texttt{CC18}     \\
1565  & \texttt{heart-h}                                                           & 294     & 14    & 5  & \texttt{TZ}       \\
1590  & \texttt{adult}                                                             & 48,842  & 15    & 2  & \texttt{CC18}     \\
1596  & \texttt{covertype}                                                         & 581,012 & 55    & 7  & \texttt{TL}       \\
4134  & \texttt{Bioresponse}                                                       & 3,751   & 1,777 & 2  & \texttt{CC18,TZ}  \\
4534  & \texttt{PhishingWebsites}                                                  & 11,055  & 31    & 2  & \texttt{CC18}     \\
4538  & \texttt{GesturePhaseSegmentationProcessed}                                 & 9,873   & 33    & 5  & \texttt{CC18,TZ}  \\
6332  & \texttt{cylinder-bands}                                                    & 540     & 40    & 2  & \texttt{CC18}     \\
23381 & \texttt{dresses-sales}                                                     & 500     & 13    & 2  & \texttt{CC18}     \\
23512 & \texttt{higgs}                                                             & 98,050  & 29    & 2  & \texttt{TZ}       \\
23517 & \texttt{numerai28.6}                                                       & 96,320  & 22    & 2  & \texttt{CC18}     \\
40536 & \texttt{SpeedDating}                                                       & 8,378   & 121   & 2  & \texttt{TL}       \\
40646 & \texttt{GAMETES\_Epistasis\_2-Way\_20atts\_0.1H\_EDM-1\_1}                  & 1,600   & 21    & 2  & \texttt{TL}       \\
40647 & \texttt{GAMETES\_Epistasis\_2-Way\_20atts\_0.4H\_EDM-1\_1}                  & 1,600   & 21    & 2  & \texttt{TL}       \\
40648 & \texttt{GAMETES\_Epistasis\_3-Way\_20atts\_0.2H\_EDM-1\_1}                  & 1,600   & 21    & 2  & \texttt{TL}       \\
40649 & \texttt{GAMETES\_Heterogeneity\_20atts\_1600\_Het\_0.4\_0.2\_50\_EDM-2\_001} & 1,600   & 21    & 2  & \texttt{TL}       \\
40650 & \texttt{GAMETES\_Heterogeneity\_20atts\_1600\_Het\_0.4\_0.2\_75\_EDM-2\_001} & 1,600   & 21    & 2  & \texttt{TL}       \\
40668 & \texttt{connect-4}                                                         & 67,557  & 43    & 3  & \texttt{CC18}     \\
40670 & \texttt{dna}                                                               & 3,186   & 181   & 3  & \texttt{CC18}     \\
40680 & \texttt{mofn-3-7-10}                                                       & 1,324   & 11    & 2  & \texttt{TL}       \\
40681 & \texttt{mux6}                                                              & 128     & 7     & 2  & \texttt{TL}       \\
40682 & \texttt{thyroid-new}                                                       & 215     & 6     & 3  & \texttt{TL}       \\
40685 & \texttt{shuttle}                                                           & 58,000  & 10    & 7  & \texttt{TL}       \\
40701 & \texttt{churn}                                                             & 5,000   & 21    & 2  & \texttt{CC18}     \\
40900 & \texttt{Satellite}                                                         & 5,100   & 37    & 2  & \texttt{TL}       \\
40945 & \texttt{Titanic}                                                           & 1,309   & 14    & 2  & \texttt{TL}       \\
40966 & \texttt{MiceProtein}                                                       & 1,080   & 82    & 8  & \texttt{CC18}     \\
40975 & \texttt{car}                                                               & 1,728   & 7     & 4  & \texttt{CC18}     \\
40978 & \texttt{Internet-Advertisements}                                           & 3,279   & 1,559 & 2  & \texttt{CC18}     \\
40979 & \texttt{mfeat-pixel}                                                       & 2,000   & 241   & 10 & \texttt{CC18}     \\
40981 & \texttt{Australian}                                                        & 690     & 15    & 2  & \texttt{TZ}       \\
40982 & \texttt{steel-plates-fault}                                                & 1,941   & 28    & 7  & \texttt{CC18}     \\
40983 & \texttt{wilt}                                                              & 4,839   & 6     & 2  & \texttt{CC18}     \\
40984 & \texttt{segment}                                                           & 2,310   & 20    & 7  & \texttt{CC18}     \\
40994 & \texttt{climate-model-simulation-crashes}                                  & 540     & 21    & 2  & \texttt{CC18}     \\
41027 & \texttt{jungle\_chess\_2pcs\_raw\_endgame\_complete}                        & 44,819  & 7     & 3  & \texttt{CC18,TZ}  \\
41138 & \texttt{APSFailure}                                                        & 76,000  & 171   & 2  & \texttt{TL}       \\
41143 & \texttt{jasmine}                                                           & 2,984   & 145   & 2  & \texttt{TZ}       \\
41147 & \texttt{albert}                                                            & 425,240 & 79    & 2  & \texttt{TZ}       \\
41150 & \texttt{MiniBooNE}                                                         & 130,064 & 51    & 2  & \texttt{TZ}       \\
43945 & \texttt{electricity}                                                       & 38,474  & 9     & 2  & \texttt{TZ}       \\
43973 & \texttt{phoneme}                                                           & 3,172   & 6     & 2  & \texttt{TZ}       \\
46905 & \texttt{Amazon\_employee\_access}                                          & 32,769  & 10    & 2  & \texttt{TA}       \\
46906 & \texttt{anneal}                                                            & 898     & 39    & 5  & \texttt{TA}       \\
46908 & \texttt{APSFailure}                                                        & 76,000  & 171   & 2  & \texttt{TA}       \\
46910 & \texttt{bank-marketing}                                                    & 45,211  & 14    & 2  & \texttt{TA}       \\
46911 & \texttt{Bank\_Customer\_Churn}                                             & 10,000  & 11    & 2  & \texttt{TA}       \\
46912 & \texttt{Bioresponse}                                                       & 3,751   & 1,777 & 2  & \texttt{TA}       \\
46913 & \texttt{blood-transfusion-service-center}                                  & 748     & 5     & 2  & \texttt{TA}       \\
46915 & \texttt{churn}                                                             & 5,000   & 20    & 2  & \texttt{TA}       \\
46916 & \texttt{coil2000\_insurance\_policies}                                     & 9,822   & 86    & 2  & \texttt{TA}       \\
46918 & \texttt{credit-g}                                                          & 1,000   & 21    & 2  & \texttt{TA}       \\
46919 & \texttt{credit\_card\_clients\_default}                                    & 30,000  & 24    & 2  & \texttt{TA}       \\
46920 & \texttt{customer\_satisfaction\_in\_airline}                               & 129,880 & 22    & 2  & \texttt{TA}       \\
46921 & \texttt{diabetes}                                                          & 768     & 9     & 2  & \texttt{TA}       \\
46922 & \texttt{Diabetes130US}                                                     & 71,518  & 48    & 2  & \texttt{TA}       \\
46924 & \texttt{E-CommereShippingData}                                             & 10,999  & 11    & 2  & \texttt{TA}       \\
46927 & \texttt{Fitness\_Club}                                                     & 1,500   & 7     & 2  & \texttt{TA}       \\
46929 & \texttt{GiveMeSomeCredit}                                                  & 150,000 & 11    & 2  & \texttt{TA}       \\
46930 & \texttt{hazelnut-spread-contaminant-detection}                             & 2,400   & 31    & 2  & \texttt{TA}       \\
46932 & \texttt{heloc}                                                             & 10,459  & 24    & 2  & \texttt{TA}       \\
46933 & \texttt{hiva\_agnostic}                                                    & 3,845   & 1,618 & 3  & \texttt{TA}       \\
46935 & \texttt{HR\_Analytics\_Job\_Change\_of\_Data\_Scientists}                   & 19,158  & 13    & 2  & \texttt{TA}       \\
46937 & \texttt{in\_vehicle\_coupon\_recommendation}                               & 12,684  & 25    & 2  & \texttt{TA}       \\
46938 & \texttt{Is-this-a-good-customer}                                           & 1,723   & 14    & 2  & \texttt{TA}       \\
46939 & \texttt{kddcup09\_appetency}                                               & 50,000  & 213   & 2  & \texttt{TA}       \\
46940 & \texttt{Marketing\_Campaign}                                               & 2,240   & 26    & 2  & \texttt{TA}       \\
46941 & \texttt{maternal\_health\_risk}                                            & 1,014   & 7     & 3  & \texttt{TA}       \\
46947 & \texttt{online\_shoppers\_intention}                                       & 12,330  & 18    & 2  & \texttt{TA}       \\
46950 & \texttt{polish\_companies\_bankruptcy}                                     & 5,910   & 65    & 2  & \texttt{TA}       \\
46952 & \texttt{qsar-biodeg}                                                       & 1,054   & 42    & 2  & \texttt{TA}       \\
46955 & \texttt{SDSS17}                                                            & 78,053  & 12    & 3  & \texttt{TA}       \\
46956 & \texttt{seismic-bumps}                                                     & 2,584   & 16    & 2  & \texttt{TA}       \\
46958 & \texttt{splice}                                                            & 3,190   & 61    & 3  & \texttt{TA}       \\
46960 & \texttt{students\_dropout\_and\_academic\_success}                         & 4,424   & 37    & 3  & \texttt{TA}       \\
46962 & \texttt{taiwanese\_bankruptcy\_prediction}                                 & 6,819   & 95    & 2  & \texttt{TA}       \\
46963 & \texttt{website\_phishing}                                                 & 1,353   & 10    & 3  & \texttt{TA}       \\
46969 & \texttt{NATICUSdroid}                                                      & 7,491   & 87    & 2  & \texttt{TA}       \\
46979 & \texttt{jm1}                                                               & 10,885  & 22    & 2  & \texttt{TA}       \\
46980 & \texttt{MIC}                                                               & 1,699   & 112   & 8  & \texttt{TA}       \\
\end{longtable}
\endgroup

\end{document}